\documentclass[11pt]{article}

\usepackage[final]{acl}

\usepackage{times}
\usepackage{latexsym}

\usepackage[T1]{fontenc}

\usepackage[utf8]{inputenc}

\usepackage{microtype}

\usepackage{inconsolata}

\usepackage{graphicx}
\usepackage{multirow}
\usepackage{colortbl}
\usepackage{algorithm}
\usepackage{algorithmic}
\usepackage{makecell}
\usepackage{svg} 
\usepackage{booktabs}
\usepackage{amsmath} 
\usepackage{amssymb}
\usepackage{hyperref}[colorlinks,linkcolor=blue]
\title{Multi-Image Visual Token Pruning in Large Visual Language Models}

\author{
  \textbf{Rongyang Zhang\textsuperscript{1}},
  \textbf{Chengqiang Lu\textsuperscript{2}},
  \textbf{Cong Li\textsuperscript{2}},
  \textbf{Hongchao Gu\textsuperscript{1}},
\\
  \textbf{Tingjia Shen\textsuperscript{1}},
  \textbf{Xuyang Zhi\textsuperscript{1}},
  \textbf{Qimeng Wang\textsuperscript{2}},
  \textbf{Yan Gao\textsuperscript{2}},
\\
  \textbf{Yi Wu\textsuperscript{2}},
  \textbf{Yao Hu\textsuperscript{2}},
  \textbf{Hao Wang\textsuperscript{1,*}},
  \textbf{Enhong Chen\textsuperscript{1,*}}
\\
\\
  \textsuperscript{1}State Key Laboratory of Cognitive Intelligence,
  University of Science and Technology of China
\\
  \textsuperscript{2}Xiaohongshu Inc.
}

\begin{document}
\maketitle

\begin{abstract}
    
With the growing demand for processing multiple image sequences in real-world applications, various visual token pruning methods have emerged to mitigate the computational and context length constraints faced by Large Vision Language Models (LVLMs).
However, most existing pruning approaches rely on static strategies that struggle to adapt across different architectural LVLMs and multi-image scenarios, and are additionally constrained by their dependence on attention computations that are incompatible with efficient techniques like FlashAttention.
To address these limitations, we propose a training-free, Adaptive Visual Token Pruning (AVTP) framework, applicable to diverse LVLM architectures. We strategically determine pruning layers based on empirical analysis of visual attention distributions across various LVLMs, and implement adaptive pruning ratios in multi-image contexts where images of higher importance retain proportionally more tokens. 
We conduct extensive experiments across different LVLMs to demonstrate the effectiveness and robustness of AVTP. Specifically, Qwen3VL-8B achieves 2× inference speedup while maintaining 96.1\% of its original accuracy on multiple multi-image benchmarks, InternVL3.5-8B retains 94.1\% accuracy, and LLaVA-OV-7B even exceeds its original baseline performance. Our code is available at \href{https://github.com/zry13/AVTP}{this link}.

\end{abstract}

\section{Introduction}
\label{sec:intro}

Large Vision-Language Models (LVLMs) have become a research focal point, demonstrating remarkable visual understanding and reasoning capabilities \cite{liu2023llava, li2024minigeminiminingpotentialmultimodality, bai2025qwen25vltechnicalreport, zhu2025internvl3exploringadvancedtraining, kimiteam2025kimivltechnicalreport, lv2026cacheinformationextractionenhanced}. They are increasingly deployed in critical domains including autonomous driving, medical diagnosis, and robotics \cite{cui2023surveymultimodallargelanguage, zhou2025improvingmedicallargevisionlanguage, li2023manipllmembodiedmultimodallarge, kim2024openvlaopensourcevisionlanguageactionmodel, wang2025generativelargerecommendationmodels}, marking significant progress toward practical multimodal AI systems. However, LVLMs encounter significant computational challenges stemming from visual data redundancy \cite{shao2025holitomholistictokenmerging, tao2025dycokedynamiccompressiontokens, yang2024visionziplongerbetternecessary, yin2024entropylawstorydata}. Visual tokens increase quadratically with the resolution, generating sequences containing thousands of tokens \cite{liu2024llavanext}. Given the transformers' quadratic complexity scaling, computational costs become prohibitive. Therefore, eliminating redundant visual tokens while preserving critical information is essential for practical LVLM deployment.

To address this challenge, numerous visual tokens pruning methods \cite{shang2024llavaprumergeadaptivetokenreduction, liu2024multistagevisiontokendropping, zhang2025vscanrethinkingvisualtoken, zhang2025sparsevlmvisualtokensparsification, yang2025topvcompatibletokenpruning, chen2025variationawarevisiontokendropping} have emerged to eliminate redundant visual tokens, thereby enhancing LVLM efficiency while preserving performance. Existing methods primarily focus on selecting the most salient tokens based on attention scores \cite{arif2024hiredattentionguidedtokendropping, han2025filtercorrelatecompresstrainingfree, xing2025pyramiddropacceleratinglargevisionlanguage, lin2025boostingmultimodallargelanguage, ye2024fitprunefasttrainingfree}. For example, FAST-V \cite{chen2024imageworth12tokens} leverages shallow-layer attention information from the LLM backbone to measure the importance of visual tokens, retaining those deemed more significant, and VisionZip \cite{yang2024visionziplongerbetternecessary} utilizes attention weights from the final layer of the Vision Transformer (ViT) to remove redundant visual tokens. Other methods like LLaVA-mini \cite{zhang2025llavaminiefficientimagevideo} employ a pre-fusion approach \cite{li2023blip2bootstrappinglanguageimagepretraining} to represent visual information using minimal tokens, and DivPrune \cite{alvar2025divprunediversitybasedvisualtoken} adopts a clustering-based method to merge similar visual tokens.
Despite their effectiveness, these approaches still suffer from notable limitations across three critical dimensions: mechanistically, they employ attention-based pruning strategies that are incompatible with efficient attention methods like FlashAttention; architecturally, they demonstrate limited generalizability across different LVLM designs; and contextually, they are primarily optimized for single-image scenarios rather than complex multi-image tasks, thereby limiting their applicability and personalization \cite{lv2026costeercollaborativedecodingtimepersonalization, lv2026specsteersynergizinglocalcontext} in real-world scenarios.

Specifically, (1) Mechanism incompatibility: The reliance on attention computations conflicts with attention acceleration frameworks such as FlashAttention, significantly diminishing the practical value of these pruning methods, as directly applying FlashAttention to uncompressed models often yields better overall performance than using pruning techniques that cannot leverage such optimizations. (2) Poor model generalizability: Prior work demonstrates that LLaVA 
models exhibit high visual token attention in shallow layers, leading to commonly adopted shallow-layer pruning approaches. However, this pattern does not generalize to other LVLM architectures, as our experimental analysis reveals that visual attention may concentrate at different layer depths, resulting in poor performance when existing methods are transferred to other models. (3) Context limitations: Existing methods are typically designed for single-image inputs and employ uniform pruning across all image tokens without considering the importance of individual tokens. In multi-image contexts, where images vary significantly in information content and relevance, this indiscriminate approach causes disproportionate information loss, leading to severe performance degradation after pruning.

\begin{figure*}[htbp]
\centering
\includegraphics[width=15cm]{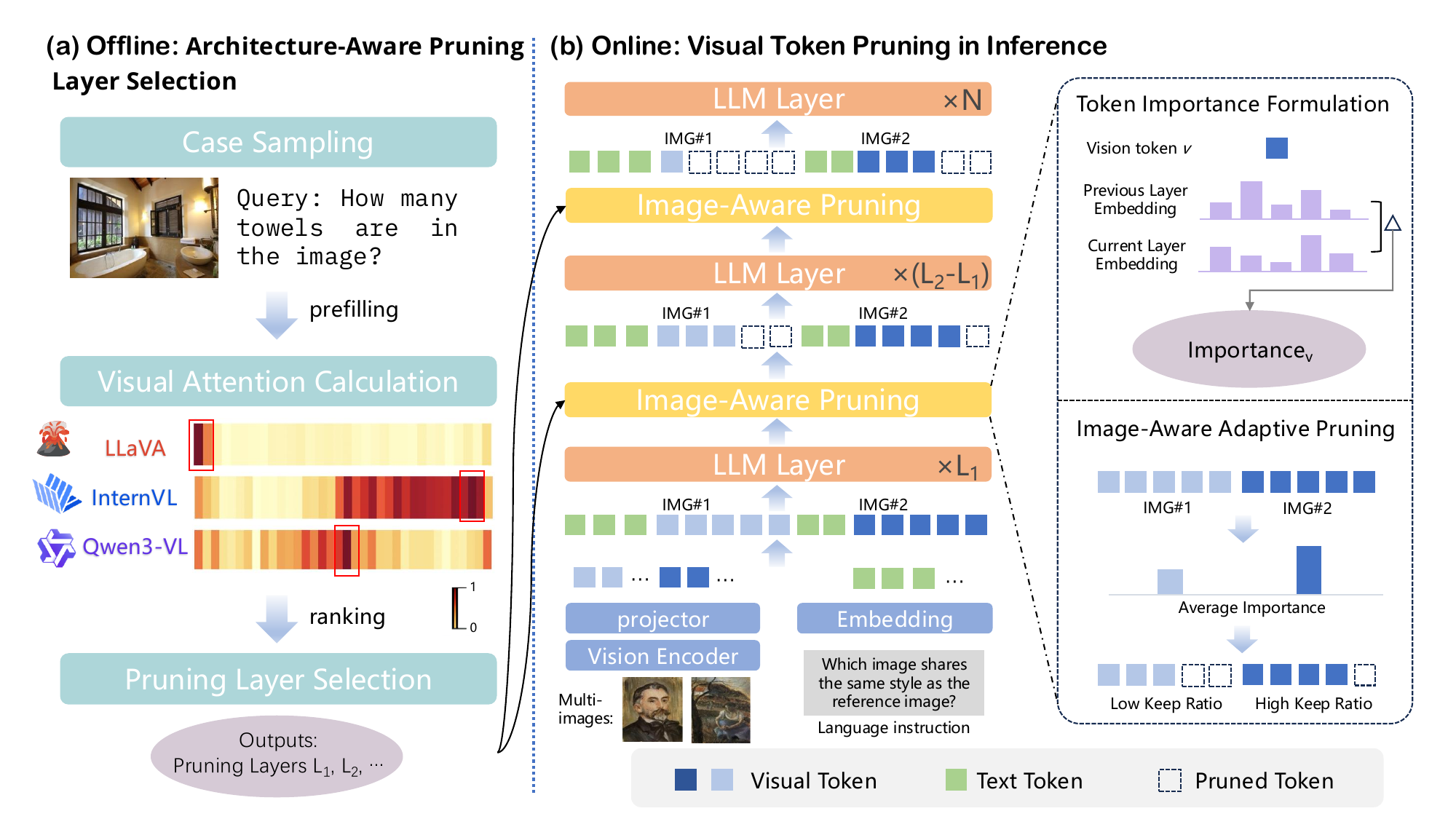}
\caption{Overall framework of our proposed AVTP. Based on varying visual attention patterns across LVLMs, we first identify high-attention layers through sampling-based analysis. During inference, we perform image-aware pruning at these layers by dynamically assigning keep ratios based on image importance.}
\label{overview}
\vspace{-0.5cm}
\end{figure*}

As shown in the Figure \ref{overview}, to address these three critical limitations and achieve efficient pruning compatible with attention acceleration methods in multi-image scenarios, we propose a unified Adaptive Visual Token Pruning (AVTP) framework with three key components that directly correspond to the identified issues. First, to tackle mechanism incompatibility, we establish a training-free pruning method based on token embedding variations across multiple layers, eliminating reliance on attention computation while providing comprehensive token importance assessment compatible with FlashAttention frameworks.
Second, to address poor model generalizability, our analysis reveals that different LVLM architectures exhibit distinct visual attention patterns—LLaVA models focus on early layers while InternVL3.5 and Qwen3-VL emphasize intermediate-to-late layers. We develop an architecture-aware pruning layer selection method to automatically adapt to these architectural differences. 
Third, to overcome context limitations, we implement image-aware adaptive pruning rates for multi-image scenarios, assigning higher pruning rates to less relevant images while preserving critical visual information from important ones. This unified approach addresses computational efficiency, architectural generalizability, and multimodal scalability simultaneously, delivering superior acceleration with maintained accuracy across diverse models and tasks. Our contributions are:
\begin{itemize}
\setlength{\itemsep}{0pt}
\item We discover distinct visual attention patterns across LVLM architectures and develop an architecture-aware layer selection method that addresses poor model generalizability.
\item We propose an image-aware adaptive pruning strategy that dynamically adjusts keep ratios based on image importance, achieving more efficient pruning in multi-image contexts.
\item We conduct extensive experiments on three mainstream LVLMs across five multi-image benchmarks, demonstrating superior performance and the effectiveness of our AVTP.

\end{itemize}

\section{Related Work}
\paragraph{Large Vision-Language Models.}
Current large vision-language models (LVLMs) integrate a vision encoder, visual projection module, and large language model to enable multi-modal comprehension. Recent work introduces higher-resolution inputs for superior performance via dynamic cropping approaches like InternVL-2.5 \cite{chen2025expandingperformanceboundariesopensource} and LLaVA-OneVision \cite{li2024llavaonevisioneasyvisualtask}, as well as native-resolution methods such as Qwen2-VL and \cite{wang2024qwen2vlenhancingvisionlanguagemodels} and Seed1.5-VL \cite{guo2025seed15vltechnicalreport}. Similarly, video large language models (VideoLLMs) process increasingly longer sequences, as demonstrated by LLaVA-Video \cite{zhang2025llavavideovideoinstructiontuning} and VideoLLaMA3 \cite{zhang2025videollama3frontiermultimodal}, with VideoXL-Pro 
\cite{liu2025videoxlproreconstructivetokencompression} achieves multi-hour frame-level understanding. However, this evolution substantially increases visual tokens, introducing quadratic complexity that severely limits LVLM scalability. 
\paragraph{Token Pruning for LVLMs.}
Token pruning methods for LVLMs can be broadly classified into four categories \cite{shao2025tokenstalkmuchsurvey} based on their core mechanisms: transformation-based approaches \cite{liu2025lacoefficientlayerwisecompression, yao2024decodecouplingtokencompression, li2023llamavidimageworth2, wang2024dynamicvlmsimpledynamicvisual} that compress tokens by directly modifying their scale or internal representation to reduce computational overhead; similarity-based techniques \cite{bolya2023tokenmergingvitfaster, wang2025dymudynamicmergingvirtual, alvar2025divprunediversitybasedvisualtoken, cao2023pumerpruningmergingtokens} that identify and eliminate redundant tokens by analyzing inherent resemblances and correlations between multimodal elements; attention-based strategies \cite{chen2024imageworth12tokens, yang2024visionziplongerbetternecessary, yang2025vflowopttokenpruningframework, zhuang2024st3acceleratingmultimodallarge} that exploit the natural sparsity patterns within attention mechanisms to selectively compress less important tokens; and query-based methods \cite{zhang2025llavaminiefficientimagevideo, alayrac2022flamingovisuallanguagemodel, song2024moresimpleeffectivetoken, zhu2023minigpt4enhancingvisionlanguageunderstanding} that leverage prompt guidance to strategically identify and filter out the most irrelevant tokens while preserving task-critical information. However, most of these methods perform pruning at shallow layers or before the LLM backbone, failing to account for the distinct characteristics of different LVLMs. Additionally, they treat all vision tokens as a unified entity, which often results in suboptimal performance in multi-image scenarios. Our proposed method addresses both limitations.

\section{Method}

\subsection{Preliminary}
\paragraph{LVLM Architecture.} Large Vision Language Models (LVLMs) typically process paired inputs denoted as ($T, V$), where T represents the textual input and $V$ denotes the visual input, such as images or videos. The textual input is encoded into $N$ textual tokens $E_t = \{t_1, ..., t_N\}$ via a text encoder. Similarly, the visual input undergoes processing through a corresponding visual encoder, which takes the visual information V as input and generates image features that are subsequently transformed into $M$ visual tokens (where $M \gg N$). These textual and visual tokens are then concatenated and fed into the Large Language Model (LLM) backbone to generate predictions through autoregressive decoding. Formally, the generation process for $O$ output tokens $Y = \{y_1, ..., y_O\}$ is formulated as follows:
\begin{center}
    $P(y_1,...,y_O|E_t,E_v)=\prod \limits_{i=1}^OP(y_i|y_{<i},E_t,E_v)$
\end{center}
where $P (|)$ is the conditional probability obtained at the output of the LLM.

\paragraph{Visual Token Pruning.} The primary objective of visual token pruning is to reduce the number of redundant visual tokens, thereby decreasing the memory footprint and inference latency of vision-language models. More formally, let $E_v$ denote the initial sequence containing $n$ projected visual tokens. The task is to select a subsequence $E_{pruned} = p(E_v)$ that can concisely yet comprehensively represent the original sequence. 

\subsection{Observations}\label{3.2}

\begin{figure}
    \centering
    \includegraphics[width=0.99\columnwidth]{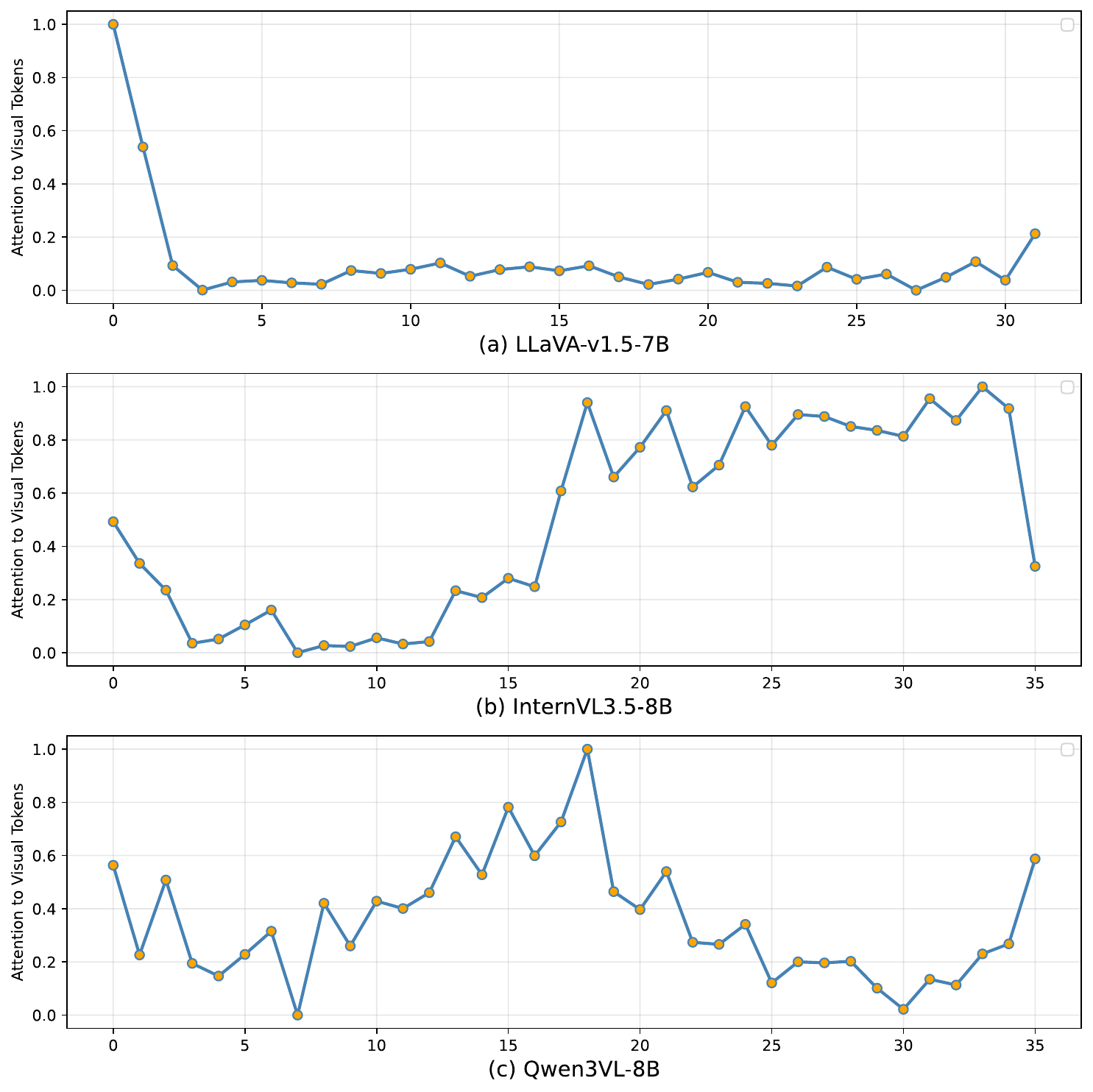}
    \caption{Attention distribution to visual tokens across layers during first token generation in different LVLMs.}
    \label{model_attn}
    \vspace{-0.5cm}
\end{figure}

\paragraph{Pruning layer selection.}
Mainstream pruning methods typically perform attention-based pruning in the shallow layers of the LLM backbone. For instance, Fast-V \cite{chen2024imageworth12tokens} retains visual tokens with higher attention values at the second layer of the LLM, based on the premise that LLMs exhibit higher overall attention distribution to visual tokens in shallow layers. However, these approaches are often limited to LLaVA-series models. We systematically evaluate the attention distribution to visual tokens across all layers of different LVLMs and find that this characteristic is not universal across all LVLMs. As illustrated in the Figure \ref{model_attn}, LLaVA-series models indeed demonstrate higher attention distribution to visual tokens in shallow layers. However, Qwen3-VL and InternVL-3.5 exhibit different patterns, with these models showing equally high attention distribution to visual tokens in middle and deeper layers. Furthermore, we observe that this distribution pattern remains relatively stable for a given model—regardless of input variation, the layers that exhibit higher attention distribution to visual tokens are consistently the same. Therefore, we propose an architecture-aware layer selection strategy for pruning, rather than being confined to shallow layers.

\paragraph{Limitation in multi-image scene.} Flash-attention \cite{dao2023flashattention2fasterattentionbetter} does not return intermediate attention values. Consequently, attention-based pruning methods require additional computation and storage to maintain flash-attention compatibility, paradoxically reducing efficiency. The extra storage requirements cause frequent out-of-memory (OOM) errors in multi-image scenarios. As shown in Table \ref{OOM_statistics}, testing mainstream pruning methods on InternVL-3.5-8B \cite{wang2025internvl35advancingopensourcemultimodal} across multi-image benchmarks using a single A100 GPU reveals frequent OOM occurrences without flash-attention, especially on MuirBench \cite{wang2024muirbenchcomprehensivebenchmarkrobust} and MIRB \cite{zhao2024benchmarkingmultiimageunderstandingvision} with high visual token counts ($>8k$). While DivPrune \cite{alvar2025divprunediversitybasedvisualtoken} maintains flash-attention compatibility through similarity-based pruning, it shows significant performance drops on multi-image benchmarks by treating all visual tokens uniformly. Therefore, a flash-attention compatible pruning method tailored for multi-image scenarios is urgently needed.

\begin{table}[!h]
    \caption{OOM case counts of existing vision token pruning methods on InternVL3.5-8B in multi-image benchmarks. Experiments were performed on a single A100 80GB GPU.}
    \label{OOM_statistics}
    \centering
    \scalebox{0.7}{
    \begin{tabular}{l|ccc}
        \toprule
        Models & Muirbench & MIRB & BLINK \\
        \midrule
        SDPA Attention & 248 & 238 & 0 \\
        Flash Attention \cite{dao2023flashattention2fasterattentionbetter} & 0 & 0 & 0 \\
        FAST-V \cite{chen2024imageworth12tokens} & 1070 & 397 & 81 \\
        VisionZip \cite{yang2024visionziplongerbetternecessary} & 1632 & 617 & 605 \\
        \makecell[l]{DivPrune(w/SDPA)\\ \cite{alvar2025divprunediversitybasedvisualtoken}}  & 1451 & 620 & 695 \\
        \bottomrule
    \end{tabular}}
    \vspace{-0.5cm}
\end{table}
\subsection{Our proposed method}
In this section, we present our pruning methodology comprising three components. First, we introduce a token importance assessment method based on hidden state variations across layers. Then, we describe an architecture-aware layer selection strategy that identifies optimal pruning layers based on visual token attention distributions within the LLM backbone. Finally, we present image-aware adaptive pruning that assigns different rates to images according to their importance.

\subsubsection{Token Importance Formulation.}\label{3.3.1}
To achieve compatibility with efficient attention mechanisms like FlashAttention and eliminate the memory overhead of storing intermediate attention weights, we leverage the magnitude of hidden state variations of visual tokens during forward propagation through the LLM backbone to assess token importance. The method requires only forward passes and maintains full compatibility with FlashAttention. The underlying insight is that important tokens contribute more substantial gradients to the final output, thus exhibiting greater variations throughout the forward propagation process.

Specifically, for a given visual token $v_i$ at layer $l$, we denote $h^{l}_{v_i}$ as its hidden state at the $l$-th layer and $h^{prior}_{v_i}$as its hidden state from the prior layer. We define the importance score $I_{v_i}$ as:
\begin{center}
    $I_{v_i} =  1-Sim(h_{v_i}^{l}, h_{v_i}^{prior})$
\end{center}
In this context, the "prior layer" denotes the immediately preceding pruning layer, with $h^{prior}_{v_i}$ representing the token embedding obtained after the most recent pruning operation. For the initial pruning layer, $h^{prior}_{v_i}$ corresponds to the original token embedding before any pruning interventions.
By evaluating tokens based on their cumulative transformations throughout the forward propagation process, our approach identifies tokens that consistently contribute meaningful information across the network. This strategy captures the global significance of visual tokens, as their importance should be determined by sustained relevance and contribution to the model's representational capacity across multiple computational stages, ultimately leading to more accurate and effective pruning decisions. 


\subsubsection{Architecture-Aware Pruning Layer Selection.}\label{3.3.2}
To address the poor model generalizability issue where existing pruning methods fail to adapt across different LVLM architectures, we develop an architecture-aware pruning layer selection strategy. As discussed in Section \ref{3.2}, while LVLMs exhibit consistent patterns in visual token attention distribution across layers within each architecture, these attention-critical layers vary systematically across different model designs.

To identify these critical layers, for a given LVLM with $L$ layers, we sample $N$ data instances $\{D_1,D_2,...,D_N\}$ and perform prefilling while recording the cumulative attention values that each layer $l\in\{1,2,...,L\}$ assigns to all visual tokens. For each sample $D_i$, we compute the attention score $A^{(i)}_l$ for layer $l$ as:
\begin{center}
    $A^{(i)}_l=\sum^M_{j=1}Attention(l, v^{(i)}_j)$
\end{center}
Where $M$ is the number of visual tokens in sample $D_i$, and $v^{(i)}_j$ represents the $j$-th visual token. Based on these attention values, we rank all layers for each sample and compute the average ranking across all samples:
\begin{center}
    $R_l=\frac{1}{N}\sum^N_{i=1}Rank(A^{(i)}_l)$
\end{center}
We then select the top-k layers with the highest average attention scores as pruning layers and implement our pruning strategy on selected layers with a progressive pruning schedule, ensuring optimal performance across diverse LVLM architectures.


\subsubsection{Image-Aware Adaptive Pruning.}

To address the context limitations where existing methods employ uniform pruning across all images without considering individual importance, we develop an image-aware adaptive pruning strategy tailored for multi-image scenarios. In multi-image contexts, images exhibit varying degrees of relevance to a given query, with some images containing more task-relevant information while others contribute marginally. This heterogeneity necessitates a differentiated approach to token preservation: images with higher importance should retain more visual tokens, whereas less important images can accommodate more aggressive pruning.

For a multi-image input with $K$ images, we compute the average importance score for image $i$ as:
\begin{center}
    $\overline{I}_i=\frac{1}{|V_i|}\sum_{v_j\in V_i}I_{v_j}$
\end{center}
where $V_i$ represents the visual tokens of image $i$. We then assign adaptive keep ratios based on relative importance:
\begin{center}
    $r_i =r_{base}+\alpha \cdot(\overline{I}_i-\overline{I}_{avg})$
\end{center}
where $r_base$ is the base keep ratio, $\alpha$ controls the variance in pruning rates, and $\overline{I}_{avg}$ is the average importance across all images. This ensures that images with higher token importance receive elevated keep ratios, while less important images undergo more aggressive pruning. 

This adaptive strategy enables fine-grained resource optimization in multi-image contexts, allocating computational resources preferentially to the most informative visual content while maintaining overall efficiency. Combined with our token importance formulation and architecture-aware layer selection, these three components systematically address the mechanism incompatibility, poor model generalizability, and context limitations of existing pruning methods, forming a unified framework that ensures both computational efficiency and performance preservation across diverse LVLM architectures and application scenarios.

\begin{table*}[!h]
    \caption{Performance comparison of various token pruning methods across different model architectures and multi-image benchmarks, with \textbf{best} results highlighted. In the latency part, “-” indicates that accurate measurement is not possible due to out-of-memory (OOM) cases.}
    \label{main_result}
    \centering
    \scalebox{0.7}{
    \begin{tabular}{l|lcccccccccc}
        \toprule
        \multirow{2}*{Model} & \multirow{2}*{Method} & \multicolumn{2}{c}{MuirBench} & \multicolumn{2}{c}{MIRB} & \multicolumn{2}{c}{BLINK} & \multicolumn{2}{c}{Qbench2} & \multicolumn{2}{c}{NLVR2}\\
        \cmidrule(r){3-4} \cmidrule(r){5-6} \cmidrule(r){7-8} \cmidrule(r){9-10} \cmidrule(r){11-12} ~& & Acc$\uparrow$ & Latency$\downarrow$ & Acc$\uparrow$ & Latency$\downarrow$ & Acc$\uparrow$ & Latency$\downarrow$ & Acc$\uparrow$ & Latency$\downarrow$ & Acc$\uparrow$ & Latency$\downarrow$ \\
        \midrule
        \rowcolor{gray!20}
        InternVL3.5-8B & Base & 52.31 & 2638.22 & 45.51 & 1317.56 & 47.45 & 900.23 & 57.90 & 566.56 & 84.83 & 3764.01 \\
         & +FAST-V & 30.23 & - & 26.75 & - & 43.87 & - & \textbf{58.00} & - & 84.17 & - \\
         & +VisionZip & 18.46 & - & 15.99 & - & 30.46 & - & \textbf{58.00} & - & 48.80 & - \\
         & +DivPrune & 43.49 & 1909.11 & 34.83 & 952.35 & 41.03 & 821.28 & 45.80 & 483.63 & 79.00 & 4139.23 \\
         & +V2DROP & 49.31 & 1597.78 & 37.32 & 769.09 & 45.23 & 613.20 & 57.00 & 371.44 & 80.72 & 2522.77 \\
         & +AVTP(Ours) & \textbf{50.38} & 1587.38 & \textbf{39.68} & 748.43 & \textbf{46.77} & 606.74 & 57.80 & 367.94 & \textbf{84.44} & 2357.15\\
         \midrule
         \rowcolor{gray!20}
        LLaVA-OneVision-7B & Base & 31.92 & 5315.54 & 21.40 & 1602.20 & 44.87 & 1951.26 & 45.20 & 1080.90 & 88.37 & 6336.99\\
         & +VisionZip & 32.31 & - & 21.79 & - & 42.56 & - & 48.90 & - & 78.02 & -\\
         & +DivPrune & 36.42 & 4616.83 & 24.61 & 1299.67 & 42.61 & 1625.67 & 51.5 & 892.64 & 84.73 & 5484.9\\
         & +V2DROP & \textbf{36.54} & 2679.29 & 24.28 & 960.53 & 43.98 & 1182.01 &  50.00 & 741.92 & 88.37 & 3837.14 \\
         & +AVTP(Ours) & 36.50 & 2680.76 & \textbf{24.86} & 958.44 & \textbf{44.87} & 1188.51 & \textbf{52.20} & 747.18 & \textbf{88.53} & 3863.38 \\
         \midrule
         \rowcolor{gray!20}
        Qwen3VL-8B & Base & 64.81 & 2086.79 & 45.21 & 875.92 & 56.71 & 1149.3 & 67.00 & 1560.66 & 87.99 & 4267.04\\
         & +FAST-V & 55.15 & - & 39.59 & - & \textbf{56.60} & - & 41.10 & - & 79.45 & -\\
         & +DivPrune & 64.04 & 1799.38 & 40.77 & 772.85 & 52.34 & 824.19 & 65.00 & 1327.65 & 84.79 & 3604.06\\
         & +V2DROP & 63.69 & 1348.66 & 36.23 & 637.96 & 50.50 & 676.77 & 63.90 & 1132.73 & 84.05 & 2696.68\\
         & +AVTP(Ours) & \textbf{65.54} & 1454.00 & \textbf{41.88} & 684.91 & 54.55 & 700.32 & \textbf{65.10} & 1184.74 & \textbf{86.15} & 2794.22\\
        \bottomrule
    \end{tabular}}
\end{table*}

\section{Experiments}

\subsection{Experimental Setting}
To demonstrate the effectiveness of our method, we conduct mainly experiments on InternVL3.5-8B \cite{wang2025internvl35advancingopensourcemultimodal}, LLaVA-ov-7B \cite{li2024llavaonevisioneasyvisualtask}, and Qwen3VL-8B \cite{bai2025qwen3vltechnicalreport}. InternVL3.5-8B is a production-oriented model that couples an 8B InternLM backbone with a dynamic-resolution ViT and 128 K context, jointly pre-trained for high-res OCR and long-document reasoning. LLaVA-OV-7B is a research-friendly open LVLM that fuses a 7B Vicuna backbone with an AnyRes ViT, jointly trained on 85M open-source image-text pairs for zero-shot transfer across single-image, multi-image, and video tasks. Qwen3-VL-8B is a recently open-sourced, state-of-the-art multimodal model that couples an 8B-parameter Qwen3 backbone with a DeepStack ViT and a 256K-token context window. It is jointly trained for high-resolution OCR, long-video event localization, and GUI agent tasks.

We conduct comprehensive experiments on five multi-image benchmark datasets: MuirBench \cite{wang2024muirbenchcomprehensivebenchmarkrobust}, MIRB \cite{zhao2024benchmarkingmultiimageunderstandingvision}, BLINK \cite{fu2024blinkmultimodallargelanguage}, Qbench2 \cite{zhang2024qbenchbenchmarkmultimodalfoundation}, and NLVR2 \cite{suhr2017corpus}. Most cases in these benchmarks involve multiple images. In particular, MuirBench comprises 2,600 cases with an average of 4.3 images per case, MIRB contains 1,013 cases with an average of 3.78 images per case, BLINK includes 1,901 cases with an average of 1.98 images per case, Q-Bench2 encompasses 1,000 cases with an average of 2.0 images per case, and NLVR2 contains 6,967 cases with an average of 2.0 images per case. We compare our approach against four state-of-the-art baseline methods: FAST-V \cite{chen2024imageworth12tokens}, VisionZip \cite{yang2024visionziplongerbetternecessary}, DivPrune \cite{alvar2025divprunediversitybasedvisualtoken}, and V2DROP \cite{chen2025variationawarevisiontokendropping}. For the base setting, we implement Flash Attention 2 \cite{dao2023flashattention2fasterattentionbetter} as the comparative baseline. For FAST-V, we rank visual tokens based on the attention weights from output tokens to visual tokens at the second layer of the LLM. For VisionZip, we utilize the attention weights from the [CLS] token in the final layer of the ViT to rank visual tokens. For DivPrune, we perform visual token pruning prior to their input into the LLM. For V2DROP, we conduct pruning at layers 3, 17, and 22 of the LLM. In our proposed AVTP, we implement Dynamic Pruning Layer Selection. To determine the optimal pruning layers, we sampled 100 cases each from four datasets: MMBench \cite{liu2024mmbenchmultimodalmodelallaround}, POPE \cite{li2023evaluating}, ScienceQA \cite{lu2022learn}, and MMStar \cite{chen2024we} for layer selection analysis and $alpha$ is set to 1.0. All methods ultimately retain 50\% of the visual tokens, and all experiments were conducted on a single A100 GPU.

\subsection{Main result}
As demonstrated in the table \ref{main_result}, we conducted comprehensive comparisons of different token pruning methods across two distinct models and five multi-image benchmarks. Selection of pruning layers was determined prior to formal experimentation using the methodology described in Section \ref{3.3.2}. In particular, for the LLaVA-onevision model, we selected only the first layer for pruning, for InternVL3.5-8B, we identified layers 1, 19, and 25 as the optimal pruning layers, and for Qwen3-VL-8B, we chose layers 1, 14, and 19. FAST-V and VisionZip encountered out-of-memory (OOM) issues on several benchmarks, which led to reduced accuracy and made reliable latency measurement impossible. In addition, LLaVA-OneVision-7B fails to produce meaningful outputs when combined with FAST-V, and Qwen3-VL-8B employs a merger module to compress ViT visual features, making it incompatible with VisionZip. Therefore, we did not conduct experiments for these two settings. Through comprehensive analysis of the experimental results, we derived several key insights:

\textbf{(i)The necessity of visual token pruning in multi-image scenarios.} In multi-image scenarios, the number of input images increases the overall sequence length substantially. Under such conditions, pruning visual tokens not only accelerates model inference but can also improve performance in certain cases.
As shown in the table \ref{main_result}, our pruning framework consistently reduces inference latency across all evaluated models. Moreover, after pruning, LLaVA-OneVision-7B achieves better performance on most multi-image benchmarks compared to its non-pruned counterpart. We attribute this to the fact that, without pruning, the input sequence length exceeds the context length the model was exposed to during training, which leads to degraded performance. Pruning shortens the sequence back into the model’s “comfort zone” of context length, thereby yielding improved results.

\textbf{(ii) Critical importance of compatibility with attention acceleration mechanisms in multi-image scenarios.} In multi-image scenarios, models are required to process very long input sequences. Performing a global sort over attention scores after they are computed incurs substantial memory overhead, which can easily lead to out-of-memory (OOM) failures. Our experiments with VisionZip and Fast-V corroborate this issue: both methods exhibit pronounced memory bottlenecks when applied to long-sequence, multi-image inputs. Moreover, when compared to a baseline that only employs FlashAttention2, VisionZip and Fast-V do not offer any advantage in inference efficiency, which severely limits their practicality in real-world deployments.
In contrast, our proposed AVTP is fully compatible with FlashAttention2 and does not introduce significant additional memory consumption. By accepting a slight degradation in performance metrics, AVTP yields substantial improvements in inference throughput and latency. This enables efficient and scalable deployment in large-scale multi-image settings while maintaining competitive model performance.

\textbf{(iii) Superior performance of our AVTP method.} As shown in Table \ref{main_result}, integrating AVTP consistently improves or matches the best accuracy while maintaining competitive latency. For InternVL3-8B, AVTP achieves the highest or near-highest accuracy on MuirBench, BLINK, Qbench2, and NLVR2, and does so with latency comparable to or lower than other compression methods such as FAST-V, VisionZip, DivPrune, and V2DROP. A similar trend holds for LLaVA-OneVision-7B, where AVTP delivers the best overall accuracy on most benchmarks, without incurring the substantial latency overhead observed in approaches that rely on expensive attention reordering. For Qwen3VL-8B, AVTP again attains the strongest results on MuirBench and Qbench2, reinforcing that its performance gains are consistent rather than model-specific. Overall, AVTP offers a more favorable accuracy–latency trade-off than prior methods, which may occasionally perform well on individual tasks but lack robustness across benchmarks and settings.

\textbf{(iv) Consistent acceleration across different model architectures.} The experiments cover three representative LVLM families—InternVL3-8B, LLaVA-OneVision-7B, and Qwen3VL-8B—which differ in backbone design, vision–language fusion mechanisms, and pretraining data. Across all these architectures, AVTP can be seamlessly integrated and yields consistent benefits, indicating that the framework is largely architecture-agnostic. In contrast, some competing methods exhibit noticeable sensitivity to the underlying model, with gains that diminish or even disappear when transferred from one LVLM to another. The stable improvements brought by AVTP across heterogeneous models demonstrate its robustness and practicality as a unified acceleration and pruning framework for large vision-language models.

These findings collectively highlight the practical advantages of our approach in real-world multi-image processing scenarios, where memory efficiency and computational acceleration are paramount considerations.

\subsection{Ablation Study}

\paragraph{Impact of Different Components in AVTP.} 
To validate dynamic layer selection and image-aware adaptive pruning, we conducted comprehensive ablation experiments. For scenarios without layer selection, we used default layers 3, 17, and 22, consistent with the V2DROP implementation. 
As shown in Table \ref{ablation1}, removing both dynamic layer selection (LS) and adaptive pruning rates (AP) from InternVL3.5-8B and Qwen3VL-8B causes substantial performance degradation across MuirBench, MIRB, and BLINK. Reintroducing either component individually yields consistent accuracy improvements, with AP showing particularly pronounced gains on BLINK. The complete AVTP configuration achieves optimal performance across all benchmarks, demonstrating strong synergy between LS and AP.
Specifically, LS dynamically identifies pruning layers, while AP calibrates pruning intensity based on image content to align with sample-specific redundancy patterns. This ablation study validates that these components drive AVTP's superior performance.

\begin{table}[!h]
    \caption{Impact of Different Components in AVTP. “LS” denotes that Dynamic Pruning Layer Selection, while “AP” indicates that Image-Aware Adaptive Pruning Rates.}
    \label{ablation1}
    \centering
    \scalebox{0.7}{
    \begin{tabular}{ccccc}
        \toprule
        Model & Method & Muirbench & MIRB & BLINK \\
        \midrule
        InternVL3.5-8B & w/o LS\&AP & 49.31 & 37.32 & 45.23 \\
        InternVL3.5-8B & w/o LS & 50.12 & 38.13 & 45.88\\
        InternVL3.5-8B & w/o AP & 49.58 & 38.41 & 46.35\\
        \rowcolor{gray!20}
        InternVL3.5-8B & AVTP & 50.38 & 39.68 & 46.77\\
        \midrule
        Qwen3VL-8B & w/o LS\&AP & 63.69 & 36.23 & 50.50\\
        Qwen3VL-8B & w/o LS & 64.05 & 37.34 & 51.39\\
        Qwen3VL-8B & w/o AP & 64.96 & 38.87 & 53.18\\
        \rowcolor{gray!20}
        Qwen3VL-8B & AVTP & 65.54 & 39.88 & 54.55\\
        \bottomrule
    \end{tabular}}
\end{table}

\paragraph{Performance Analysis with Variable Image Quantities.}
To further validate the effectiveness of AVTP in multi-image scenarios, we analyzed the effect of different token pruning methods on Qwen3VL-8B under varying numbers of input images on the Muirbench. As shown in the Table \ref{ablation2}, our proposed AVTP method consistently outperforms other approaches across varying input image quantities. AVTP achieves the highest scores in three out of four configurations, demonstrating superior performance compared to the original model. While FAST-V suffers significant performance drops (especially 37.38 for $\ge 8$ images), and V2DROP shows good stability, AVTP maintains robust performance even with larger image sets, establishing its effectiveness for multi-image vision-language tasks.

\begin{table}[!h]
    \caption{Effect of Different Pruning Methods on Qwen3VL-8B Performance Across Varying Numbers of Input Images.}
    \label{ablation2}
    \centering
    \scalebox{0.7}{
    \begin{tabular}{cccccc}
        \toprule
        Image Num & Original & FAST-V & DivPrune & V2DROP & AVTP \\
        \midrule
        2-3 & 60.32 & 57.34 & 59.70 & 59.32 & \textbf{61.19} \\
        4-5 & 67.73 & 57.61 & 66.44 & 66.13 & \textbf{68.49}\\
        6-7 & 67.39 & 50.36 & 68.48 & 68.12 & \textbf{68.84}\\
        $\ge 8$ & 60.19 & 37.38 & 59.71 & 59.22 & \textbf{60.12}\\
        \bottomrule
    \end{tabular}}
\end{table}

\paragraph{Analysis of Different Pruning Layer Effects.}

\begin{figure}
    \centering
    \includegraphics[width=0.99\columnwidth]{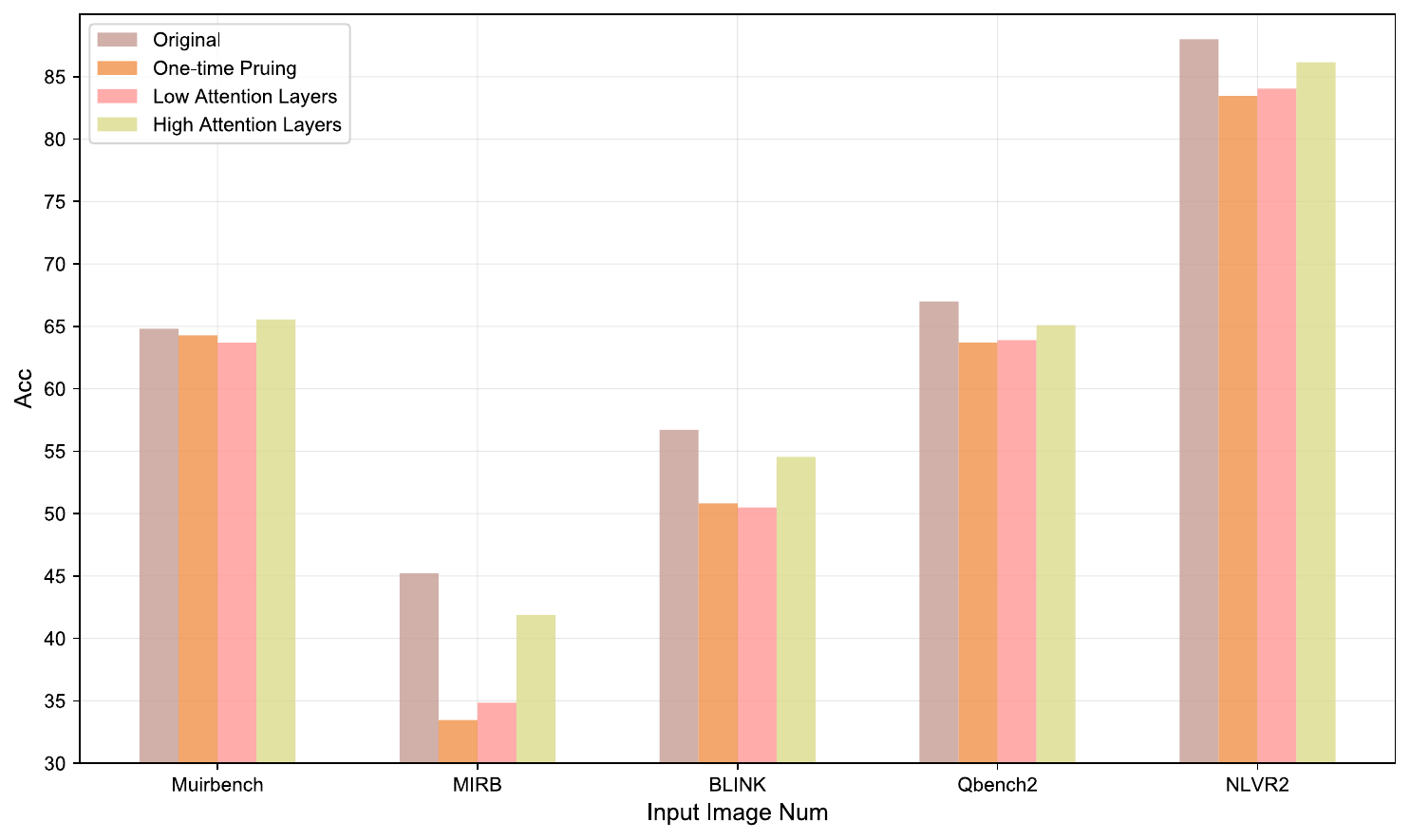}
    \caption{Performance of Different Pruning Layers on Qwen3VL-8B.}
    \label{ablation3}
    \vspace{-0.5cm}
\end{figure}

To evaluate the effectiveness of our layer selection strategy, we conduct comparative experiments on Qwen3VL-8B under three pruning configurations: (1) pruning only Layer 1; (2) pruning Layers 1, 17, and 26, where the latter two exhibit relatively low visual attention; and (3) applying AVTP to prune Layers 1, 14, and 19, which exhibit high visual attention. As shown in Figure~\ref{ablation3}, pruning only Layer 1 results in the lowest performance on MIRB, QBench2, and NLVR2. Similarly, pruning layers with low visual attention leads to inferior results on MuirBench and BLINK. Conversely, pruning high-attention layers consistently achieves optimal results, substantially outperforming other strategies. These results validate the effectiveness of our method and highlight the importance of selecting layers with strong visual attention.


\section{Conclusion}

In this work, we address two key limitations of existing vision token pruning methods for multimodal large language models: their dependence on specific model architectures and their suboptimal performance in multi-image scenarios. To this end, we propose Adaptive Visual Token Pruning (AVTP), a training-free framework that adaptively selects pruning layers across diverse architectures through Dynamic Pruning Layer Selection. AVTP estimates visual token importance based on hidden-state variations, supported by theoretical analysis, and further introduces Image-Aware Adaptive Pruning Rates to allocate pruning budgets across multiple images. Extensive experiments on a range of multi-image benchmarks demonstrate that AVTP consistently outperforms existing visual token pruning methods. Moreover, ablation studies validate the effectiveness of each component and highlight their complementary benefits.

\section{Limitations}
Despite the promising results, our work has several limitations that warrant discussion. 

First, several critical hyperparameters within the AVTP framework suffer from insufficient theoretical grounding and are presently established through exhaustive empirical experimentation. In particular, the top-k parameter employed in Architecture-Aware Pruning Layer Selection, the scaling factor $\alpha$ governing keep ratio variations in Image-Aware Adaptive Pruning, and the correlation between visual token embedding modifications and token significance, though conceptually plausible, lack formal mathematical substantiation.

Second, our experimental evaluation primarily focuses on comparisons with other visual token pruning methods, without comprehensive analysis against complementary inference acceleration techniques such as quantization and knowledge distillation. It is worth noting that our approach is inherently compatible with these orthogonal techniques and can be synergistically deployed, representing different technological pathways rather than competing alternatives. 

Finally, our visual token importance formulation advances upon the established V2DROP methodology. While our approach delivers substantial performance improvements, the methodological contributions represent an evolutionary advancement rather than a fundamentally novel algorithmic paradigm.

\section*{Acknowledgments}

This work was supported by the National Natural Science Foundation of China (Nos. U23A20319, 62472394, 62441239, and 62441227), as well as the Anhui Province Science and Technology Innovation Project (Nos. 202423k09020010 and 202423k09020011).

\bibliography{custom}

\newpage
\appendix

\begin{table*}[htbp]
    \caption{Ablation Study on Different Pruning Ratios}
    \label{pruning_ratio}
    \centering
    \scalebox{0.8}{
    \begin{tabular}{l|lcccccccccc}
        \toprule
        \multirow{2}*{Model} & \multirow{2}*{Keep Ratio} & \multicolumn{2}{c}{MuirBench} & \multicolumn{2}{c}{MIRB} & \multicolumn{2}{c}{BLINK} & \multicolumn{2}{c}{Qbench2} & \multicolumn{2}{c}{NLVR2}\\
        \cmidrule(r){3-4} \cmidrule(r){5-6} \cmidrule(r){7-8} \cmidrule(r){9-10} \cmidrule(r){11-12} ~& & Acc$\uparrow$ & Latency$\downarrow$ & Acc$\uparrow$ & Latency$\downarrow$ & Acc$\uparrow$ & Latency$\downarrow$ & Acc$\uparrow$ & Latency$\downarrow$ & Acc$\uparrow$ & Latency$\downarrow$ \\
        \midrule
        Qwen3VL-8B & Base & 64.81 & 2086.79 & 45.21 & 875.92 & 56.71 & 1149.3 & 67.00 & 1560.66 & 87.99 & 4267.04 \\
         & 50\% & 65.54 & 1454.00 & 41.88 & 684.91 & 54.55 & 700.32 & 65.10 & 1184.74 & 86.15 & 2794.22\\
         & 20\% & 65.00 & 1403.39 & 37.61 & 768.69 & 54.02 & 720.66 & 63.90 & 1163.92 & 85.79 & 2842.69\\
         & 5\% & 64.50 & 1366.28 & 33.17 & 760.08 & 50.13 & 714.53 & 64.00 & 1150.51 & 85.80 & 2778.36\\
        \bottomrule
    \end{tabular}}
\end{table*}

\section{Additional Ablation Experiments}
In this section, we present comprehensive ablation studies to further validate our approach. Specifically, we systematically evaluate the effectiveness of AVTP on the Qwen3VL-8B model across various pruning ratios, and investigate the impact of the hyperparameter $\alpha$ in our Image-Aware Adaptive Pruning mechanism, which controls the variance in pruning ratios across different images. All experiments were conducted on a single A100 80GB GPU.

\subsection{Impact of Different Pruning Rates}
As demonstrated in the table \ref{pruning_ratio}, beyond our main experiments and baseline model, we further evaluate the effectiveness of our method at keeping ratios of 20\% and 5\%. The results reveal that while higher pruning ratios yield improved inference efficiency, they come with a corresponding trade-off in performance. Notably, when the keep ratio is reduced to 5\%, the Qwen3VL model exhibits substantial performance degradation: on MIRB, the score drops from the baseline's 45.21\% to 33.17\%, and on BLINK, it decreases from 56.71\% to 50.13\%. In comparison, the performance at a 50\% keep ratio achieves 41.88\% and 65.10\%, respectively, highlighting the more pronounced performance decline at the 5\% retention level.

\subsection{Effect of the Hyperparameter $\alpha$ in Image-Aware Adaptive Pruning}

In this section, we conduct an ablation study on the hyperparameter $\alpha$ in Image-Aware Adaptive Pruning. While $\alpha$ was set to 1.0 in the main experiments, we explore the effect of increasing this coefficient to amplify the influence of Image-Aware Adaptive Pruning on pruning rate allocation. It is worth noting that larger $\alpha$ values may lead to pruning rates exceeding valid bounds (i.e., greater than 1 or less than 0). To prevent such extreme cases, we impose constraints on the pruning rates for individual images, setting the maximum and minimum values at 95\% and 5\%, respectively.

As shown in the Table \ref{ablation alpha}, the experimental results reveal varying impacts across different benchmarks. For benchmarks with a larger number of input images, such as Muirbench and MIRB, increasing $\alpha$ yields slight performance improvements. However, for benchmarks with fewer input images, including BLINK, QBench2, and NLVR2, the benefits of larger $\alpha$ values are marginal, and in some cases, smaller $\alpha$ values demonstrate superior performance.

\begin{table}[!h]
    \caption{Ablation Study on Hyperparameter $\alpha$ in Image-Aware Adaptive Pruning.}
    \label{ablation alpha}
    \centering
    \scalebox{0.7}{
    \begin{tabular}{lccccc}
        \toprule
        $\alpha$ & Muirbench & MIRB & BLINK & QBench2 & NLVR2 \\
        \midrule
         1.0 & 65.54 & 41.88 & 54.55 & 65.10 & 86.15 \\
         2.0 & 65.27 & 41.95 & 55.18 & 64.70 & 85.80 \\
         3.0 & 65.08 & 42.15 & 55.02 & 64.90 & 86.39 \\
         5.0 & 65.65 & 42.56 & 55.02 & 64.70 & 86.42 \\
        \bottomrule
    \end{tabular}}
\end{table}

\begin{table*}[htbp]
    \caption{Performance on Single-Image and Video Benchmarks, “-” indicates that accurate measurement is not possible due to out-of-memory (OOM) cases.  }
    \label{other bench}
    \centering
    \scalebox{0.75}{
    \begin{tabular}{l|lcccccccccc}
        \toprule
        \multirow{2}*{Model} & \multirow{2}*{Method} & \multicolumn{2}{c}{MMBench} & \multicolumn{2}{c}{MMStar} & \multicolumn{2}{c}{POPE} & \multicolumn{2}{c}{MME} & \multicolumn{2}{c}{MVBench}\\
        \cmidrule(r){3-4} \cmidrule(r){5-6} \cmidrule(r){7-8} \cmidrule(r){9-10} \cmidrule(r){11-12} ~& & Acc$\uparrow$ & Latency$\downarrow$ & Acc$\uparrow$ & Latency$\downarrow$ & Acc$\uparrow$ & Latency$\downarrow$ & Score$\uparrow$ & Latency$\downarrow$ & Acc$\uparrow$ & Latency$\downarrow$ \\
        \midrule
        \rowcolor{gray!20}
        Qwen3VL-8B & base & 82.55 & 620.50 & 60.40 & 257.97 & 89.03 & 1150.95 & 2219.97 & 1226.64 & 61.53 &  1239.18 \\
         & +FAST-V & 79.23 & 467.92 & 55.33 & 180.84 & 88.70 & 744.53 & 2163.58 & 718.63 & 45.95 & - \\
         & +DivPrune & 81.49 & 495.48 & 53.20 & 198.24 & 89.18 & 902.18 & 2130.12 & 948.46 & 61.03 & 1034.84 \\
         & +V2DROP & 81.49 & 462.08 & 57.06 & 170.86 & 88.08 & 770.89 & 2174.38 & 650.13 & 60.78 & 839.42 \\
         & +AVTP(Ours) & 81.77 & 452.42 & 57.13 & 172.19 & 88.11 & 774.11 & 2200.66 & 656.09 & 61.13 & 843.04 \\
        \bottomrule
    \end{tabular}}
\end{table*}

\section{Performance on Single-Image and Video Benchmarks}

In the main experiments and ablation studies, we primarily evaluated the effectiveness of AVTP on multi-image benchmarks. In this section, we further demonstrate the performance of AVTP on several single-image VQA benchmarks and video benchmarks. Consistent with the main experiments, all pruning methods are configured with a uniform pruning rate of 50\%, meaning that 50\% of vision tokens are retained while the remaining tokens are pruned. All other experimental settings remain identical to those described in the main text, and experiments are conducted on a single A100-80GB GPU. It is important to note that we employ custom-constructed prompts for evaluation, which may result in performance variations compared to official benchmark reports. Nevertheless, we ensure fair comparison by maintaining identical prompts across all methods throughout our evaluation.

\subsection{Evaluation Benchmark}

\paragraph{MMBench.} MMBench evaluates models through three hierarchical levels of abilities: L-1 with two core abilities (perception and reasoning), L-2 with six sub-abilities, and L-3 with 20 specific dimensions. This structure enables a detailed assessment of diverse capabilities.

\paragraph{ScienceQA.} Spanning domains like natural, language, and social sciences, ScienceQA organizes questions hierarchically into 26 topics, 127 categories, and 379 skills. This benchmark evaluates multimodal understanding, multi-step reasoning, and interpretability.

\paragraph{POPE.} POPE evaluates Object Hallucination in models using binary questions on object presence in images. Metrics such as Accuracy, Recall, Precision, and F1 Score measure hallucination levels across three sampling strategies, providing precise assessments.

\paragraph{MME.} The MME benchmark evaluates model performance across 14 subtasks targeting perceptual and cognitive abilities. Using manually designed instruction-answer pairs, MME minimizes data leakage for fair assessment.


\paragraph{MVBench.} MVBench defines 20 video understanding tasks that require deep comprehension of temporal dimensions, beyond single-frame analysis.

\subsection{Experimental Results}

We conducted additional comparative experiments on MMBench, MMStar, POPE, MME, and MVBench, as presented in the Table \ref{other bench}. When applying FAST-V on MVBench, numerous out-of-memory (OOM) cases occurred, resulting in significantly reduced accuracy and unmeasurable latency for FAST-V on this benchmark. Although our Image-Aware Adaptive Pruning does not provide benefits for single-image and video benchmarks, our AVTP method still achieves the best performance on MMBench, MMStar, MME, and MVBench. These results demonstrate the critical importance of effective pruning layer selection in optimizing model performance across diverse multimodal tasks.



\end{document}